# Toward Multimodal Conversational AI for Age-Related Macular Degeneration

**Authors**

Ran Gu[1*], Benjamin Hou[1*], Mélanie Hébert[2], Asmita Indurkar[2], Yifan Yang[1], Emily Y. Chew[2], Tiarnán D. L. Keenan[2], Zhiyong Lu[1]

**Institutions**

1. Division of Intramural Research, National Library of Medicine, National Institutes of Health, Bethesda, Maryland, USA

2. Division of Epidemiology and Clinical Applications, National Eye Institute, National Institutes of Health, Bethesda, Maryland, USA

* Contributed equally

**Corresponding Author**

Zhiyong Lu, PhD, FACMI

Division of Intramural Research,

National Library of Medicine, National Institutes of Health

8600 Rockville Pike, Bethesda, MD 20894

Telephone: 301 594 7089

Email: zhiyong.lu@nih.gov

**Abbreviations and Acronyms**

AMD, age-related macular degeneration; CFP, color fundus photograph; CLIP, contrastive language-image pre-training; MLLM, multimodal large language model; VQA, visual question answering; AMD, age-related macular degeneration; AREDS, Age-Related Eye Disease Study; CNN, convolutional neural network; DR, diabetic retinopathy; IRB, Institutional Review Board; VLM, vision-language model; OcularChat, ocular chatbot; OCT, optical coherence tomography

**Abstract**

Despite strong performance of deep learning models in retinal disease detection, most systems produce static predictions without clinical reasoning or interactive explanation. Recent advances in multimodal large language models (MLLMs) integrate diagnostic predictions with clinically meaningful dialogue to support clinical decision-making and patient counseling. In this study, OcularChat, an MLLM, was fine-tuned from Qwen2.5-VL using simulated patient-physician dialogues to diagnose age-related macular degeneration (AMD) through visual question answering on color fundus photographs (CFPs). A total of 705,850 simulated dialogues paired with 46,167 CFPs were generated to train OcularChat to identify key AMD features and produce reasoned predictions. OcularChat demonstrated strong classification performance in AREDS, achieving accuracies of 0.954, 0.849, and 0.678 for the three diagnostic tasks: advanced AMD, pigmentary abnormalities, and drusen size, significantly outperforming existing MLLMs. On AREDS2, OcularChat remained the top-performing method on all tasks. Across three independent ophthalmologist graders, OcularChat achieved higher mean scores than a strong baseline model for advanced AMD (3.503 vs. 2.833), pigmentary abnormalities (3.272 vs. 2.828), drusen size (3.064 vs. 2.433), and overall impression (2.978 vs. 2.464) on a 5-point clinical grading rubric. Beyond strong objective performance in AMD severity classification, OcularChat demonstrated the ability to provide diagnostic reasoning, clinically relevant explanations, and interactive dialogue, with high performance in subjective ophthalmologist evaluation. These findings suggest that MLLMs may enable accurate, interpretable, and clinically useful image-based diagnosis and classification of AMD.

**Keywords**

color fundus photographs, multimodal large language models, visual question answering, age-related macular degeneration

**Introduction**

Age-related macular degeneration (AMD) remains a leading cause of irreversible vision loss among older adults. It affects nearly 200 million people worldwide, with numbers projected to increase with population aging[1]. Clinically, AMD encompasses a spectrum of progressive macular changes, ranging from earlier findings, particularly drusen accumulation and pigmentary abnormalities, to advanced disease, specifically geographic atrophy and neovascular AMD[2-4]. Early identification of AMD features and accurate risk stratification are essential for patient counseling, monitoring, and timely intervention.

In clinical practice and screening, traditional diagnosis relies on human expert interpretation of retinal images within clinical context. However, this process is time-consuming, subject to inter-observer variability, and difficult to scale across large populations. Timely identification and risk stratification of AMD therefore remain challenging, highlighting the need for automated tools capable of identifying advanced disease and analyzing precursor features such as drusen characteristics and pigmentary abnormalities[5].

Automated methods, particularly those based on deep learning, have become increasingly central to medical image analysis[6-8], and have demonstrated robust performance in ophthalmology, including the detection of diabetic retinopathy (DR)[9-12], glaucoma[13-16], cataract[17-19], and AMD[5, 20-24]. Many of these systems have achieved high diagnostic accuracy for specific tasks. However, most operate as task-specific classifiers that produce categorical outputs without explanation and with limited transparency. In contrast, clinical interpretation of retinal findings is inherently contextual and communicative: clinicians must explain imaging findings, address patient questions, discuss prognosis and modifiable risk factors, and integrate imaging results into broader care decisions.

Recent advances in multimodal large language models (MLLMs), which integrate both image and language processing capabilities[25-27], have created new opportunities for clinical support. These models can answer and explain questions. Such capabilities align closely with the communicative and interpretive aspects of ophthalmic care, where effective explanation of retinal findings is central to patient counseling and shared decision-making. Rather than functioning solely as automated classifiers, multimodal systems may support clinicians by enhancing interpretability, consistency, and efficiency in retinal image assessment and communication[28-33].

In this study, we developed and evaluated OcularChat, an MLLM specifically designed for AMD-related retinal image assessment and clinically grounded dialogue. In contrast to conventional vision-language models (VLMs) that rely on contrastive language-image pre-training (CLIP)-style pretraining[28] or separate visual and language encoders[30], OcularChat was fine-tuned end-to-end from a pretrained Qwen2.5-VL model using retinal images paired with simulated patient-physician dialogues derived from the Age-Related Eye Disease Study (AREDS). We assessed its performance on three clinically relevant tasks: identifying advanced AMD, categorizing drusen size, and detecting pigmentary abnormalities. More importantly, we introduced a clinician-designed grading rubric and conducted human expert assessment to examine the quality of its explanations and conversational responses and compared these outcomes with those of the reference model.

By jointly integrating image-based diagnosis with interpretable, dialogue-based outputs and clinician-centered evaluation, this work advances beyond prior benchmarking studies of multimodal ophthalmic models toward a clinically oriented assessment framework[34]. As MLLMs begin to enter clinical practice, rigorous clinician-centered evaluation is essential to ensure not only accurate feature identification but also reliable diagnostic reasoning and dialogue that are safe and trustworthy in clinical settings.

**Results**

*Patient Sample*

The flow diagram of participant inclusion is shown in Supplementary Figure S1. A total of 59,333 CFPs, from 4,107 AREDS participants, were included in this study. For model development and evaluation, 46,167 CFPs from 3,192 participants were allocated to the training set, and 13,166 CFPs from 915 participants were allocated to the test set. As listed in Table 1, the final analyzed cohort (test set) contained 1,781 (13.53%) CFPs with advanced AMD, 4,891 (37.15%) with pigmentary abnormalities, and 4,380 (33.27%) with large drusen; mean age of the study population was 69.52 ± 5.07 years, while 501 (54.75%) participants were female. For external validation, the test set contained 3,872 CFPs from 831 participants in AREDS2 cohort, including 343 (8.86%) CFPs with advanced AMD, 3,120 (80.58%) with pigmentary abnormalities, and 3,614 (93.34%) with large drusen; mean age was 73.45 ± 7.38 years, and 495 (59.57%) were female.

Table 1. Number of color fundus photographs in the training and testing sets, categorized by macular features and advanced age-related macular degeneration status.

|  | Number of CFPs (% of Total) | | |
|  | Training | Testing | Total |
| --- | --- | --- | --- |
| **Advanced AMD** |  |  |  |
| Yes | 5,712 (12.37) | 1,781 (13.53) | 7,493 (12.63) |
| No | 40,455 (87.63) | 11,385 (86.47) | 51,840 (87.37) |
| **Pigmentary abnormalities** |  |  |  |
| Yes | 16,920 (36.65) | 4,891 (37.15) | 21,811 (36.76) |
| No | 29,247 (63.35) | 8,275 (62.85) | 37,522 (63.24) |
| **Drusen size** |  |  |  |
| Small/none | 18,960 (41.07) | 5,140 (39.04) | 24,100 (40.62) |
| Intermediate | 12,496 (27.07) | 3,646 (27.69) | 16,142 (27.20) |
| Large | 14,711 (31.86) | 4,380 (33.27) | 19,091 (32.18) |
| **Total images** | 46,167 (100.0) | 13,166 (100.0) | 59,333 (100.0) |

Abbreviations: CFPs = color fundus photographs; AMD = age-related macular degeneration

*Quantitative Evaluation of Model Accuracy Compared with Pretrained VLMs*

OcularChat generated predictions of advanced AMD, pigmentary abnormalities, and drusen size for all eyes in the test set. Quantitative results for these three tasks are summarized in Table 2. For advanced

AMD prediction, OcularChat demonstrated higher performance than both general-purpose and medically adapted VLMs. It achieved accuracy of 0.954 and F1 score of 0.850. In comparison, GPT-o1 and MedGemma-4B achieved F1 scores of 0.748 and 0.597, respectively.

Table 2. Comparisons of predicting advanced age-related macular degeneration, pigmentary abnormalities, and drusen size. 95% confidence intervals were computed using bootstrap resampling (n = 2000) with the Clopper-Pearson exact method. The highest scores are shown in bold.

| Model | Accuracy | | | F1-score | | |
|---|---|---|---|---|---|---|
| | ADVAMD | PIG | DRUS | ADVAMD | PIG | DRUS |
| Qwen2.5-VL-7B | 0.543 (0.535-0.552)* | 0.664 (0.656-0.672)* | 0.530 (0.521-0.538)* | 0.363* | 0.193* | 0.411* |
| Qwen2.5-VL-32B | 0.722 (0.715-0.730)* | 0.749 (0.742-0.757)* | 0.528 (0.520-0.537)* | 0.477* | 0.673* | 0.409* |
| Qwen2.5-VL-72B | 0.825 (0.819-0.832)* | 0.785 (0.778-0.792)* | 0.523 (0.514-0.531)* | 0.568* | 0.667* | 0.414* |
| Llama-3.2-11B-Vision | 0.769 (0.762-0.777)* | 0.570 (0.562-0.579)* | 0.393 (0.384-0.401)* | 0.286* | 0.378* | 0.359* |
| Llama-3.2-90B-Vision | 0.895 (0.889-0.900)* | 0.645 (0.637-0.653)* | 0.430 (0.422-0.439)* | 0.509* | 0.457* | 0.378* |
| GPT-o1 | 0.941 (0.937-0.945)* | 0.776 (0.769-0.783)* | 0.492 (0.483-0.500)* | 0.748* | 0.614* | 0.424* |
| MedGemma-4B | 0.889 (0.883-0.894)* | 0.737 (0.730-0.745)* | 0.484 (0.476-0.493)* | 0.597* | 0.543* | 0.418* |
| **OcularChat (ours)** | **0.954 (0.950-0.957)** | **0.849 (0.843-0.855)** | **0.678 (0.670-0.686)** | **0.837** | **0.803** | **0.642** |

∗ means the p-value < .001

Abbreviations: ADVAMD = advanced age-related macular degeneration; PIG = pigmentary abnormalities; DRUS = drusen size

For pigmentary abnormalities prediction, OcularChat again demonstrated higher performance relative to general-purpose VLMs, achieving the highest accuracy of 0.849 and F1 score of 0.803. In contrast, several general-domain models exhibited imbalanced performance; for example, Qwen2.5-VL-7B showed moderate accuracy (0.664) but low F1 score (0.194). Drusen size prediction was more challenging across all models. Despite this, OcularChat still maintained higher performance than general-purpose VLMs, achieving an F1 score of 0.642, compared with 0.424 for GPT-o1, 0.418 for MedGemma-4B, and 0.409-0.414 across Qwen2.5-VL variants. Notably, OcularChat also outperformed the Qwen2.5-VL-32B baseline model, which achieved an F1 score of 0.409.

External validations on the AREDS2 test set showed that OcularChat exhibited greater generalizability, with less performance degradation than Qwen2.5-VL-32B or GPT-o1, across the three tasks. As summarized in Table 3, for advanced AMD detection, OcularChat achieved high accuracy (0.892; 95% CI, 0.881-0.901) and F1 score (0.494), comparable to GPT-o1 and substantially outperforming Qwen2.5-VL-32B, which showed markedly reduced performance. For pigmentary abnormalities, OcularChat maintained balanced accuracy (0.786; 95% CI, 0.774-0.799) and F1 score (0.867), remaining competitive with both comparator models. Notably, for drusen size classification, OcularChat achieved the highest accuracy (0.857; 95% CI, 0.847-0.868) and F1 score (0.495), whereas both Qwen2.5-VL-32B and GPT-o1 exhibited substantially lower F1 scores despite moderate accuracy.

Table 3. External evaluations of Owen2.5-VL-32B, GPT-o1, and OcularChat on AREDS2.

| Models | Accuracy | | | F1-score | | |
| --- | --- | --- | --- | --- | --- | --- |
| | ADVAMD | PIG | DRUS | ADVAMD | PIG | DRUS |
| Qwen2.5-VL-32B | 0.225 (0.212-0.238)∗ | 0.788 (0.775-0.801) | 0.829 (0.817-0.841)∗ | 0.183∗ | 0.873 | 0.333∗ |
| GPT-o1 | 0.896 (0.886-0.905) | 0.741 (0.728-0.755)∗ | 0.608 (0.593-0.624)∗ | 0.510 | 0.830∗ | 0.326∗ |
| OcularChat (ours) | 0.892 (0.881-0.901) | 0.786 (0.774-0.799) | 0.857 (0.847-0.868) | 0.494 | 0.867 | 0.495 |

∗ means the p-value < .001.

Abbreviations: ADVAMD = advanced age-related macular degeneration; PIG = pigmentary abnormalities; DRUS = drusen size

*Model Evaluation by Ophthalmologists*

Figure 1 presents an example of an interactive patient-physician dialogue generated during inference and reviewed as part of the ophthalmologist assessment. The results of the quantitative ophthalmologist evaluation across all samples (n = 120) are summarized in Table 4. Across all ophthalmologist graders,

OcularChat achieved substantially higher mean scores than the reference model for the identification of advanced AMD (3.503 vs 2.833), pigmentary abnormalities (3.272 vs 2.828), drusen size (3.064 vs 2.433), and overall impression (2.978 vs 2.464).

**Interactive dialogue during inference**

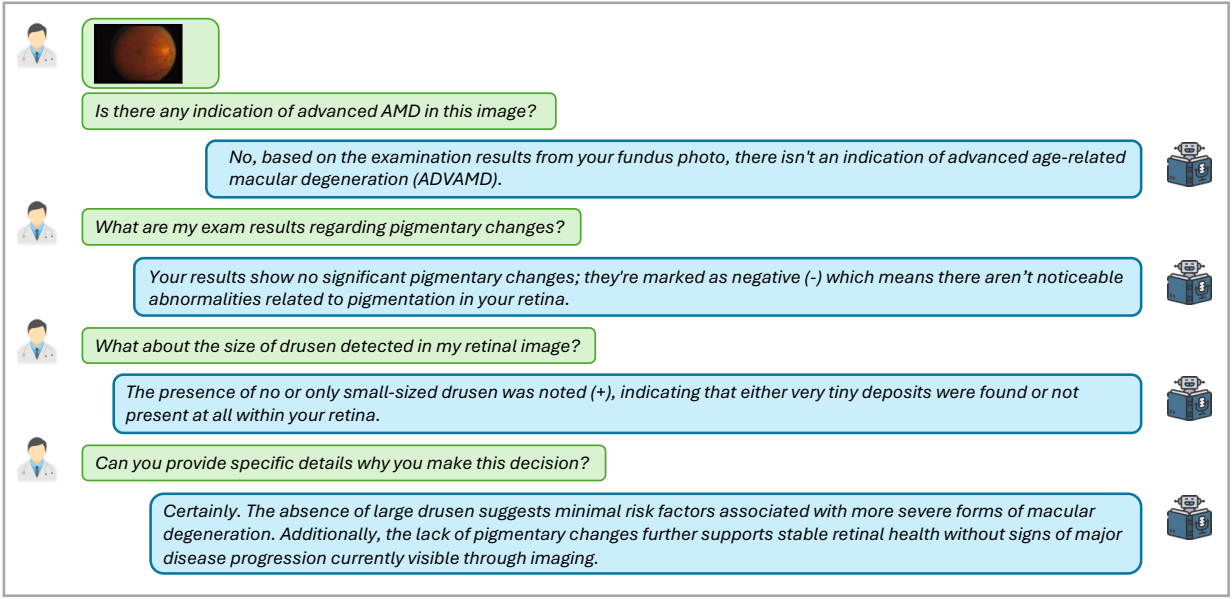

Figure 1. Example of patient-physician dialogues during inference when conducting clinical evaluation.

Table 4. Quantitative results of clinicians' evaluation of responses from OcularChat and its baseline model. Qwen is the reference model Qwen2.5-VL-32B. The highest scores are shown in bold for each question.

| Ophthalmologist | Model | ADVAMD (Q1) | PIG (Q2) | DRUS (Q3) | Overall (Q4) |
|---|---|---|---|---|---|
| 1 | Qwen | 2.967 (2.533-3.417) | 3.183 (2.783-3.583) | 3.033 (2.717-3.367) | 2.933 (2.633-3.234) |
| | OcularChat | **4.067 (3.667-4.417)** | **3.733 (3.350-4.100)** | **3.667 (3.300-4.033)** | **3.567 (3.283-3.867)** |
| 2 | Qwen | 2.950 (2.617-3.300) | 3.017 (2.683-3.333) | 2.183 (1.950-2.433) | 2.617 (2.367-2.850) |
| | OcularChat | **3.300 (2.967-3.633)** | **3.033 (2.683-3.367)** | **3.100 (2.817-3.383)** | **2.883 (2.583-3.183)** |
| 3 | Qwen | 2.583 (2.266-2.900) | 2.283 (2.017-2.567) | 2.083 (1.833-2.350) | 1.842 (1.633-2.058) |
| | OcularChat | **3.142 (2.758-3.500)** | **3.050 (2.642-3.425)** | **2.425 (2.116-2.750)** | **2.483 (2.192-2.783)** |
| Average | Qwen | 2.833 | 2.828 | 2.433 | 2.464 |
| | OcularChat | **3.503** | **3.272** | **3.064** | **2.978** |

Abbreviations: ADVAMD = advanced age-related macular degeneration; PIG = pigmentary abnormalities; DRUS = drusen size

For Ophthalmologist 1, mean scores were higher for OcularChat compared with the reference model, for example, advanced AMD (4.067 vs 2.967) and overall impression (3.567 vs 2.933). Ophthalmologist 2 demonstrated a similar pattern, with higher scores for OcularChat versus the baseline model observed for drusen size (3.100 vs 2.183) and overall impression (2.883 vs 2.617). Although Ophthalmologist 3 showed stricter grading overall, higher scores for OcularChat compared with the baseline model were also observed across all tasks, particularly for advanced AMD (3.142 vs 2.583) and overall impression (2.483 vs 1.842).

To control for case heterogeneity, a common subset of 30 samples was evaluated by ophthalmologists. Inter-rater agreement was further assessed across all four questions using Cohen's kappa. As illustrated in Figure 2, kappa heatmaps summarize agreement across all four evaluation categories and rater pairs. OcularChat demonstrated kappa values more than 20% higher than the reference model, with all inter-rater agreements exceeding 0.65. Additionally, the lower kappa values between evaluators 1 and 3, compared with the other inter-rater pairs, indicate greater variability in their evaluation preferences. Such differences highlight the need for adaptable response strategies in multimodal models, which could be explored in future work through configurable explanation depth or clinician-specific interaction modes.

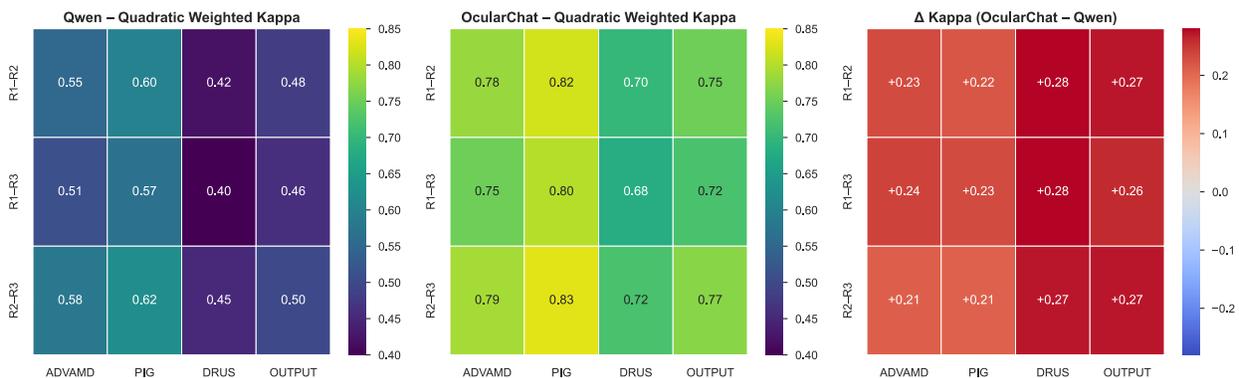

Figure 2. Inter-rater agreement for reference Qwen model and OcularChat across four clinical tasks. Kappa heatmaps show quadratic weighted kappa between the three raters (R1–R2, R1–R3, R2–R3) for Qwen model, OcularChat, and the difference map (ΔKappa = OcularChat – Qwen).

**Discussion**

In this study, OcularChat was developed as a multimodal system for retinal image interpretation, including interactive dialogue and reasoning. Across all diagnostic tasks, it achieved higher performance than general-purpose VLMs, as well as a domain-specific model. Additionally, according to subjective assessment by ophthalmologists, it achieved higher ratings for diagnostic reasoning and clinical usefulness compared with the reference model.

Most VLMs are pretrained on general-domain data, with limited exposure to ophthalmic images and terminology, which can restrict performance in detailed retinal image interpretation. OcularChat incorporates ophthalmology-specific data and dialogue-based supervision to address this gap. It was adapted from a pretrained VLM (Qwen2.5-VL-32B) and trained using CFPs and synthetic patient-physician dialogues derived from clinical information in the AREDS dataset.

External validation showed that OcularChat generalizes more consistently across tasks on external data. Across all tested models, we observed a notable reduction in performance for detecting advanced AMD in AREDS2. This decline likely reflects differences in disease appearance in the modern treatment era, in which neovascular AMD is managed with anti–vascular endothelial growth factor (VEGF) therapy and may present with subtle or absent features on CFP, in comparison to the florid appearance of untreated neovascular AMD at the time of AREDS. Nonetheless, OcularChat still maintained a high accuracy score.

Most prior works have focused on developing deep learning-based classification models that produce fixed numeric outputs, such as DeepSeeNet[5], which demonstrates strong predictive performance but provides limited interpretability. More recent efforts, including large multimodal ophthalmology dataset (LMOD)[34], evaluated VLMs for ophthalmic image understanding and diagnosis. While LMOD provides valuable insights into current VLMs' capabilities, the results highlight substantial limitations in ophthalmic image interpretation, particularly in clinically meaningful reasoning. In contrast, our work advances beyond benchmarking by developing an ophthalmology-specific VLM, with a focus on AMD-related image assessment and explanatory conversational output.

Beyond quantitative grading scores, additional insights were obtained through ophthalmologist evaluation of model outputs. This evaluation identified systematic semantic misalignments in explanatory content that were not captured by prediction accuracy alone. As illustrated in Figure 1, although OcularChat correctly identified the absence of advanced AMD, its explanation emphasized drusen size and pigmentary abnormalities rather than the defining features of advanced disease, i.e., geographic atrophy and neovascular disease. While accurate identification of all stages of AMD is clinically important, particularly for risk stratification and initiation of AREDS2 supplementation, this pattern highlights a misalignment between the diagnostic target and the explanatory focus. Specifically, conflating features of earlier disease stages with criteria for advanced AMD may obscure the distinction between disease severity categories, which is of clinical concern for neovascular AMD, where timely detection is critical given the availability of effective vision-preserving therapies. These findings indicate that correct diagnostic labels do not necessarily correspond to clinically valid or stage-appropriate reasoning.

This issue was more evident in the reference model, whose verbose outputs frequently combined some correct conclusions with inaccurate or irrelevant anatomical descriptions, including references to regions not central to the diagnostic task. Clinicians consistently noted that this response style obscured the model's final judgment. Although the reference model occasionally included appropriate guidance (e.g., recommending professional consultation or lifestyle modification), it also increased susceptibility to hallucination and knowledge bias, which was reflected in consistently lower clinician scores for relevance and internal consistency. In contrast, OcularChat produced concise, task-focused responses that reduced the likelihood of contradictory explanations and were perceived as better aligned with clinical workflows, particularly in opportunistic screening scenarios where clear identification of pathologic findings is prioritized. In practice, OcularChat is intended to function as a decision-support tool that complements clinician judgment by highlighting salient image features and providing concise explanations, rather than replacing standard diagnostic workflows.

This work has several limitations. Although OcularChat's more concise outputs were generally preferred for their efficiency and clarity, some ophthalmologists noted that highly simplified outputs insufficiently demonstrated image-grounded reasoning, raising concerns that predictions partially relied on learned priors rather than explicit visual evidence. This observation also reflects variability in clinician preferences, with some favoring concise, directive outputs and others valuing more detailed explanations to support exploratory reasoning or discussion.

Another limitation is the reliance on CFPs, as it was trained exclusively on this single image modality. Although CFPs are central to retinal evaluation, and reflect the retinal appearance on clinical examination, they do not always capture the full range of structural information required for ophthalmic diagnosis in all settings. Optical coherence tomography (OCT), in particular, provides high-resolution 3-dimensional imaging, especially for the early diagnosis of the neovascular form of late AMD[41-44]. Nevertheless, CFPs remain highly advantageous for large-scale screening applications, as they can be acquired using cost-effective imaging devices, and are widely available across diverse clinical settings.

In summary, this study demonstrates the feasibility of adapting MLLMs for retinal image interpretation, with clinician evaluation supporting their potential role as interpretable decision-support tools in AMD assessment, with scope for further expansion to broader ophthalmic applications. By integrating image-based diagnosis with clinically grounded explanations and interactive dialogue, OcularChat illustrates how MLLMs can move beyond fixed-output classification toward supporting real-world clinical reasoning and patient communication.

**Methods**

This main study used publicly available data from the AREDS[35], a multicenter 12-year clinical trial conducted across 11 clinical sites in the United States. The original AREDS study was approved by the Institutional Review Board (IRB) at all participating sites, and written informed consent was obtained from all participants. Eligible participants underwent stereoscopic color fundus photography (CFP; 30° field centered on the fovea) of both eyes at baseline and annually for up to 12 years of follow-up. Participants were categorized into AMD severity groups, including individuals with no AMD.

For the current study, CFPs and corresponding eye-level annotations (described below) were selected from the AREDS dataset. Data were split at the participant level into training, validation, and test sets to prevent information leakage, ensuring that images from the same participant did not appear in other subsets. To ensure generalizability across eyes, images from both eyes were included.

*Dataset Construction and Dialogue Generation*

To ensure clinical relevance and contextual accuracy during model training, simulated patient-physician dialogues were generated using key clinical variables extracted from the AREDS dataset. These variables included demographic characteristics (age, gender, diabetes status, and smoking history) and ophthalmic features (presence of advanced AMD, maximum drusen size, and presence of pigmentary abnormalities), defined by the AREDS simplified severity scale[35].

Dialogue generation was performed using a large language model (GPT-4V[36]), with structured prompts designed to ensure clinical coherence and to constrain dialogue content to the provided variables. The prompt design was physician-informed; clinicians were consulted to identify key questions typically asked during patient evaluation and to guide construction of the generation prompts so that simulated dialogues reflected realistic patient-physician interactions. The dialogue generation workflow is illustrated in Figure 3, and the prompts used for generation are provided in Supplementary Tables S1 and S3. The generated dialogues contained two formats intended to reflect realistic patient-physician interactions while guaranteeing the accuracy of AMD-related risk assessment (Supplementary Tables S2 and S4). A total of 705,850 patient-physician dialogues were generated through GPT-4V. Standard preprocessing followed Qwen2.5-VL recommendations[37] with details presented in Supplementary S. 1. The generated dialogues, paired with their corresponding retinal images, formed the training dataset.

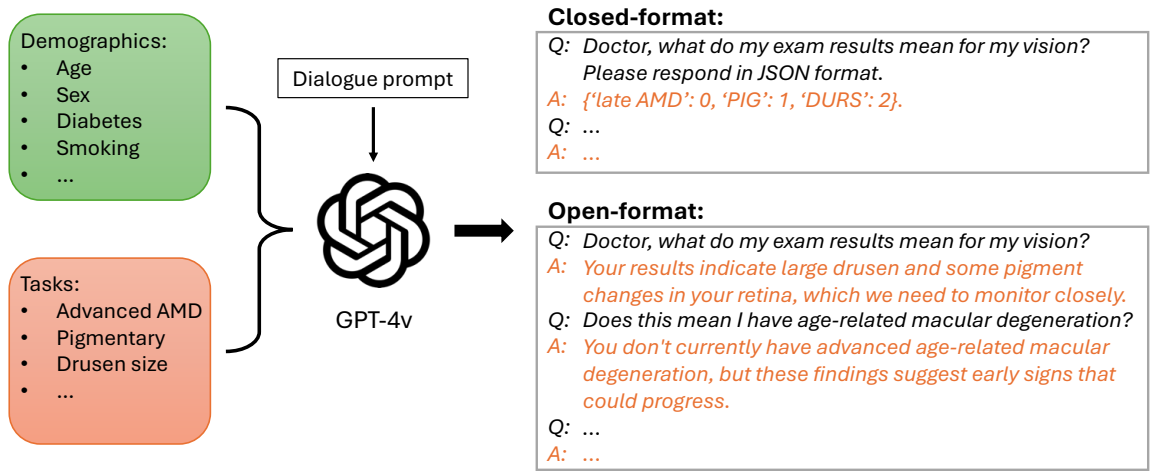

Figure 3. Description of dialogue generation strategy. Here, we generate open- and closed-format dialogues depending on GPT-4V. The key point is to find a suitable dialogue prompt to encourage GPT-4V to generate a variety of questions and answers based on the demographics and task-related information.

*Training the OcularChat Vision-Language Model*

Figure 4 illustrates the training pipeline for OcularChat, which was developed by fine-tuning a pre-trained MLLM to jointly process retinal images and dialogue-based text. The training objective aligns visual features with corresponding linguistic dialogues, enabling the model to generate clinically coherent conversations grounded in retinal imaging features. Qwen2.5-VL-32B was selected as the baseline model due to its open-source accessibility and strong multi-modal reasoning capability. It integrates a high-capacity language model with a powerful vision encoder, making it well-suited for medical applications that require image-conditioned natural language understanding and generation. All parameters of Qwen2.5-VL-32B were optimized during training; the implementations are provided in Supplementary S. 2.

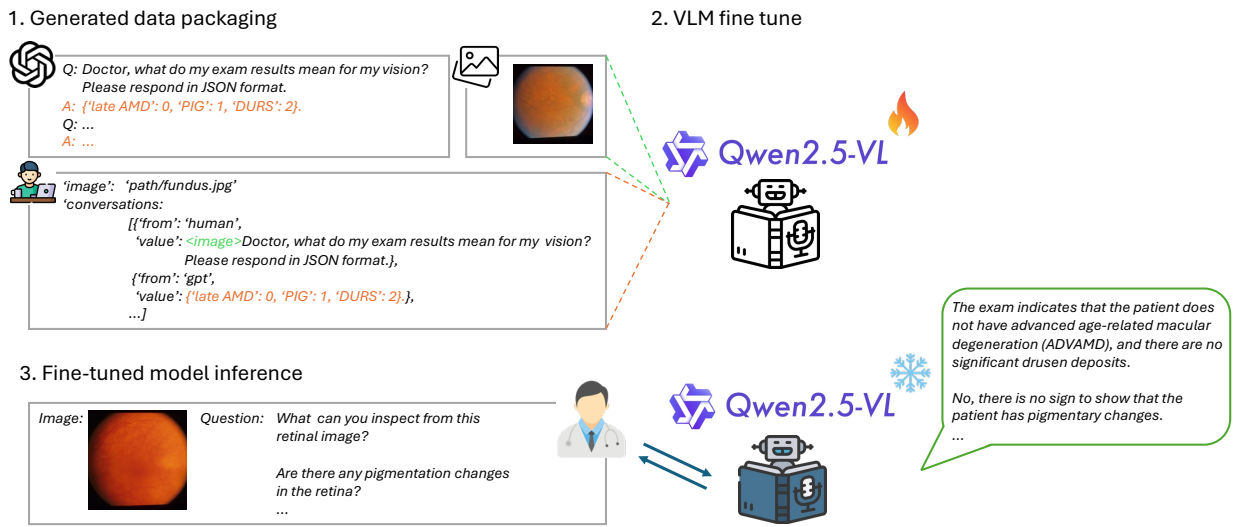

Figure 4. Framework of our proposed OcularChat for retinal image analysis, prediction, and communication online. When OcularChat is fine-tuned, users can chat with it interactively, discussing ophthalmic issues.

*Quantitative Evaluation of Model Accuracy*

For quantitative and objective evaluation of model accuracy, OcularChat was prompted to generate structured outputs for three AMD-related tasks: (1) identification of advanced AMD, (2) detection of pigmentary abnormalities, and (3) categorization of drusen size. To facilitate large-scale analysis, the model was instructed to return its outputs in structured JSON format. To evaluate performance, these outputs were compared with the ground truth labels derived from previous reading center grading of the AREDS retinal images. Specifically, performance was assessed using accuracy and F1 score. All analyses were conducted using Python, NumPy and scikit-learn (Supplementary S. 3).

The performance of OcularChat was compared with that of several state-of-the-art VLMs pre-trained on general-domain data, including the Llama-3.2-Vision[38], Qwen2.5-VL[37], and GPT-o1[39]. In addition, Med-Gemma[40], a VLM fine-tuned for medical applications, was included to provide a domain-adapted baseline model for comparison. The same prompts and decoding parameters were used to ensure fair comparison. Supplementary Table S5 details the prompts used for all evaluations.

External validation was further conducted using an independent AREDS2 cohort to assess model generalizability beyond the primary training and test data. OcularChat, its baseline model Qwen2.5-VL-32B, and GPT-o1 (that performed best among the compared models on the AREDS dataset), were evaluated using the same task definitions, prompts, output formats, and performance metrics as in the main analysis.

*Model Evaluation by Ophthalmologists*

To assess the clinical relevance and interpretability of the model outputs, human evaluation was performed by three retinal specialists. The retina specialists worked independently and were masked to model identity.

For each retinal image in the test set, interactive multi-turn dialogues were generated to ask for diagnosis and detailed explanations, using ophthalmologist-informed prompts designed to reflect questions commonly asked during real-world patient evaluation (Supplementary Table S6). These prompts were designed to elicit the model's diagnosis and reasoning for each case, to probe clinical usefulness (including responsiveness, coherence, and absence of hallucinations), and assess whether additional clinically relevant findings were detected. The ophthalmologists reviewed the dialogues generated by OcularChat and by a reference model in response to the simulated patient queries. For each dialogue, ophthalmologists had access to the dialogue, the corresponding retinal images, and the reading center grades, but were masked to the identity of the model being evaluated.

Evaluation was conducted using a custom-designed grading rubric that assessed the accuracy of the model's diagnostic reasoning, explanation quality, and clinical usefulness (Supplementary Table S7) for each of the three tasks and an overall assessment. Each task was scored on a 1-5 scale (5 = highest performance), with detailed descriptors and examples defining the criteria for each score to promote consistent grading among the ophthalmologists. The test set comprised 120 image-text pairs, drawn randomly from the larger AREDS test set while maintaining the proportions of all AMD severity levels. Of these, 30 pairs were selected as a common subset, and were evaluated independently by all three ophthalmologists, enabling estimation of inter-rater agreement. The remaining 90 pairs were partitioned into three sets of 30 pairs and were assigned to a single ophthalmologist each. Thus, each ophthalmologist evaluated 60 pairs (30 common and 30 unique). Model performance was then computed as the average rubric score among each clinician.

*Statistical Analysis*

Diagnostic performance was assessed using averaged metric scores and 95% confidence intervals were estimated using bootstrap resampling with 2,000 iterations. Confidence intervals were derived using the Clopper-Pearson exact method. Statistical significance of differences in performance between models was assessed using paired hypothesis testing, accounting for the paired nature of predictions on the same test cases. Differences in accuracy were evaluated using exact McNemar tests based on discordant prediction pairs, while differences in F1 score was assessed using paired bootstrap resampling (B = 2,000). Two-sided P values were reported. All analyses were performed at the case level unless otherwise stated.

**Data & Code Availability**

The code of this project is publicly available at: https://huggingface.co/ncbi/OcularChat.

The generated data and trained model are available at: https://huggingface.co/ncbi/OcularChat-VQA; https://huggingface.co/ncbi/OcularChat32B.

AREDS dataset: https://www.ncbi.nlm.nih.gov/projects/gap/cgi-bin/study.cgi?study_id=phs000001.v3.p1.

AREDS2 dataset: https://www.ncbi.nlm.nih.gov/projects/gap/cgi-bin/study.cgi?study_id=phs002015.v2.p1.

**Acknowledgements**

This research was supported by the Intramural Research Program of the National Institutes of Health (NIH), including the National Library of Medicine and the National Eye Institute. This work used the computational resources of the NIH high-performance computing Biowulf cluster ([http://hpc.nih.gov](http://hpc.nih.gov)). The contributions of the NIH authors were made as part of their official duties as NIH federal employees, are incompliance with agency policy requirements, and are considered Works of the United States Government. However, the findings and conclusions presented in this paper are those of the authors and do not necessarily reflect the views of the NIH or the U.S. Department of Health and Human Services.

**Ethics declarations**

Conflict of Interest: The authors have no potential conflicts of interest to disclose.

**Funding**

Funding source: Division of Intramural Research, National Library of Medicine, National Institutes of Health

Grant number: Not applicable.

**Author Contributions**

R.G., B.H., and Z.L. conceived this study. R.G. and B.H. implemented the data collection and model construction. Y.Y. provided algorithm support. R.G. and B.H. conducted model quantitative evaluation and manuscript drafting and revision. M.H, A.I., and T.K. conducted the ophthalmologist evaluation and manuscript revision. R. G. and B.H. developed the demo website of OcularChat. T.K., E.C., and Z.L. supervised the study. All authors contributed to writing the manuscript and approved the submitted version.

**Ethics Statement**

This study did not involve direct interaction with human participants or animals. All data used in this study were obtained from publicly available datasets and were de-identified prior to analysis. The study was conducted in accordance with relevant institutional guidelines and national regulations.

25. Jin Q, Chen F, Zhou Y, et al. Hidden flaws behind expert-level accuracy of multimodal GPT-4 vision in medicine. NPJ Digit Med. 2024;7(1):190. Published 2024 Jul 23. doi:10.1038/s41746-024-01185-7

26. Hamamci IE, Er S, Sekuboyina A, et al. GenerateCT: Text-conditional generation of 3D chest CT volumes. arXiv preprint arXiv:2305.16037. Published May 26, 2023. Accessed February 6, 2026. https://arxiv.org/abs/2305.16037

27. Hou B, Mukherjee P, Batheja V, Wang KC, Summers RM, Lu Z. One Year On: Assessing Progress of Multimodal Large Language Model Performance on RSNA 2024 Case of the Day Questions. Radiology. 2025;316(2):e250617. doi:10.1148/radiol.250617

28. Wang Z, Wu Z, Agarwal D, Sun J. MedCLIP: Contrastive Learning from Unpaired Medical Images and Text. Proc Conf Empir Methods Nat Lang Process. 2022;2022:3876-3887. doi:10.18653/v1/2022.emnlp-main.256

29. Moor M, Huang Q, Wu S, et al. Med-Flamingo: a Multimodal Medical Few-shot Learner. Proceedings of the 3rd Machine Learning for Health Symposium. Proceedings of Machine Learning Research. 2023;225:353-367. https://proceedings.mlr.press/v225/moor23a.html

30. Zhang K, Zhou R, Adhikarla E, et al. A generalist vision-language foundation model for diverse biomedical tasks. Nat Med. 2024;30(11):3129-3141. doi:10.1038/s41591-024-03185-2

31. Zhao Z, Liu Y, Wu H, et al. CLIP in medical imaging: A survey. Med Image Anal. 2025;102:103551. doi:10.1016/j.media.2025.103551

32. Wang W, Lv Q, Yu W, et al. CogVLM: Visual Expert for Pretrained Language Models. In: Globerson A, Mackey L, Belgrave D, et al, eds. Advances in Neural Information Processing Systems. Vol 37. NeurIPS; 2024. doi:10.52202/079017-3860. Accessed February 6, 2026. https://proceedings.neurips.cc/paper_files/paper/2024/hash/dc06d4d2792265fb5454a6092bfd5c6a-Abstract-Conference.html
25

# Supplementary

**Contents**

## S. 1 Dataset Construction and Dialogue Generation

In dialogue generation, a total of 705,850 patient-physician dialogues were generated through GPT-4V with temperature of 0.5 and 1024 maximum tokens. The generated dialogues, paired with their corresponding retinal images, formed the training dataset. Each sample consisted of a single pair of retinal images and associated dialogues formatted as alternating conversational turns. The generation strategy is shown in Figure S1. Retinal images were resized to 1024×1024 pixels and normalized following the pre-processing pipeline recommended for Qwen2.5-VL. No additional manual cropping or image enhancement was applied.

## S. 2 Training Details

All parameters of Qwen2.5-VL-32B, including the vision and language encoders and the multilayer perceptron (MLP), were optimized during training. Training was conducted using Pytorch (version 2.6) in a distributed setting using eight NVIDIA H100 GPUs, each with 80 GB of memory and per-GPU batch size of 2. An effective batch size of 128 was achieved through gradient accumulation over eight steps. Mixed-precision training with bfloat16 (BF16) was utilized to enhance memory efficiency and computational performance. The learning rate was initialized at 1e-6 and scheduled using cosine decay, with a warm-up phase comprising 3% of the total training steps to facilitate stable convergence. Optimization was performed using AdamW optimizer with default settings. The model was trained to minimize cross-entropy loss between predicted and target dialogue tokens, conditioned on both the fundus image and preceding conversational context. Model checkpoints were saved at regular intervals, and training was conducted for two epochs. Unless otherwise specified, all remaining hyperparameters followed the official recommendations for Qwen2.5-VL.

## S. 3 Evaluation of Model Accuracy

To facilitate large-scale analysis, the model was instructed to return its outputs in a structured JSON format. To evaluate performance, these outputs were compared with the ground truth labels derived from previous reading center grading of the AREDS retinal images. Specifically, performance was assessed using accuracy and F1 score. All analyses were conducted using Python (version 3.10.18), NumPy (version 2.1.2) and scikit-learn (version 1.6.1). All models were evaluated using the same prompts and decoding parameters (including temperature and maximum token length) to ensure fair comparison.

**Figure S1. Workflow of data curation from AREDS.**

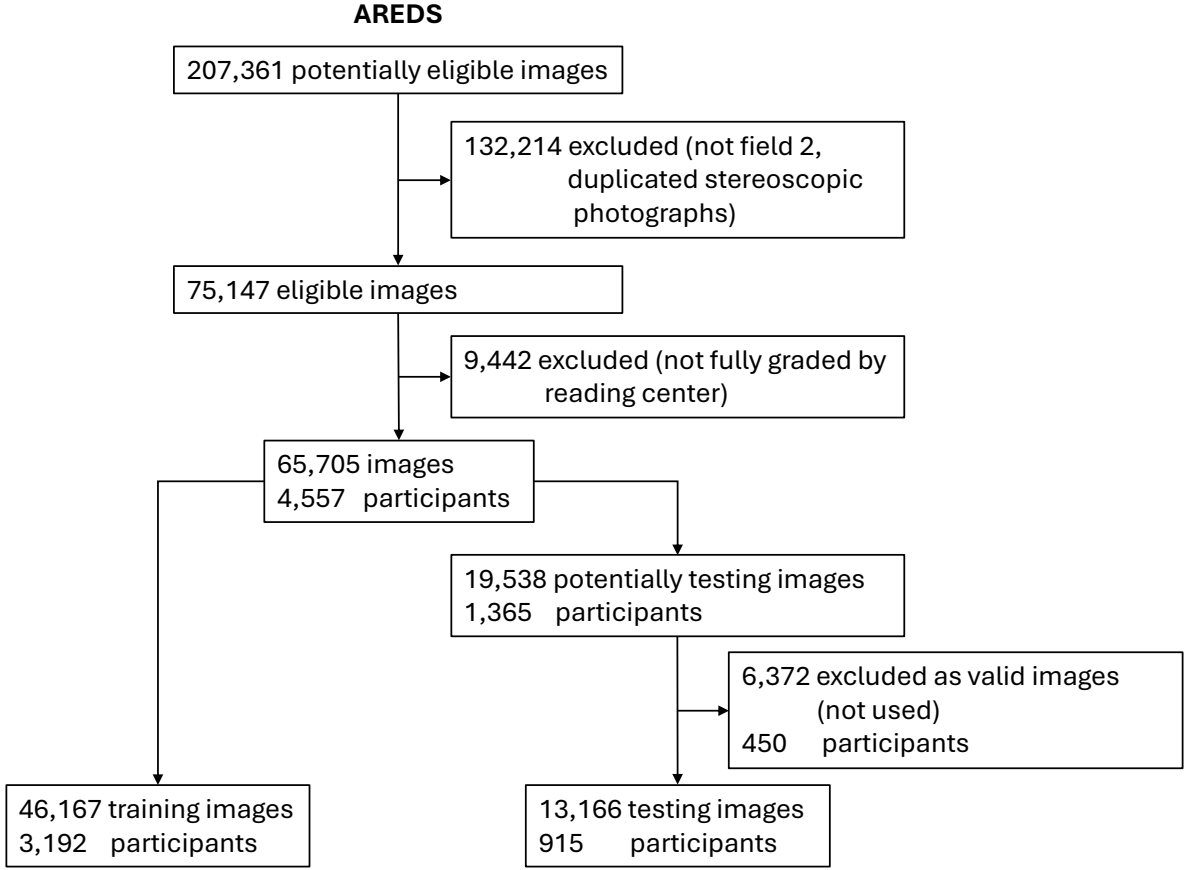

**Table S1. Open-form Dialogue Generation Prompt.**

---

You are a highly experienced ophthalmologist specializing in retinal diseases, having a conversation with a patient during a clinic visit. The patient has the following information:

- Age: {age}

- Sex: {sex} (0 = female, 1 = male)

- Diabetes: {diab} (0 = No, 1 = Yes)

- Smoker: {smk} (1 = current, 2 = former, 3 = never)

And you have patient's exam results:

- Late Age-related Macular Degeneration (advanced AMD): {lateamd} (0 = No, 1 = Yes)

- Drusen Size (DRUS): {drus} (0 = small/none, 1 = intermediate, 2 = large)

- Pigmentary (PIG): {pig} (0 = No, 1 = Yes)

Simulate a natural and concise multi-turn conversation between the patient and doctor around these topics, ensuring medical professionalism and coherence. The patient demographics information should be contained in conversation.

Respond only with a JSON array of short dialog turns in the following format:

[

{{"role": "patient", "text": "⟨question⟩"}},

{{"role": "doctor", "text": "⟨answer⟩"}},

]

Include 3 turns, each turn should be a questing and answering about the exam results. Each turn should contain one of the exam results, and 3-turn dialogues must involve all of exam result.

Avoid repeating irrelevant information, no explanation and compliment required and keep the tone professional and context-aware.

---

**Table S2. Example of Generated Open-form Dialogue.**

"conversations":

[

"from": "human", "value": "<image>What do my exam results indicate about my eyes?",

"from": "gpt", "value": "Your results show intermediate drusen but no advanced age-related macular degeneration or pigment changes.",

"from": "human", "value": "Is this something I should be concerned about given my age?",

"from": "gpt", "value": "Intermediate drusen can increase the risk of developing age-related macular degeneration, especially at your age, but there are no signs of advanced disease currently.",

"from": "human", "value": "Does smoking affect my risk of macular degeneration?",

"from": "gpt", "value": "Yes, smoking can significantly increase your risk of developing macular degeneration. It would be beneficial to consider quitting."

]

**Table S3. JSON-form Dialogue Generation Prompt.**

> You are a highly experienced ophthalmologist specializing in retinal diseases. Now you are a dialogue creator specializing in generating question-and-answer sets.
>
> You need to simulate a natural question and answering between the patient and doctor only surrounding the exam results. The patient does not know any exam results in advance and no clinical background. please ensure medical professionalism and coherence. Respond only with a JSON format of short dialog turns in the following format:
>
> [
>
> {{"role": "human", "text": "⟨question⟩"}},
>
> {{"role": "gpt", "text": "⟨answer⟩"}},
>
> ]
>
> Please create three turns question-and-answer sets, and let the question turns randomly. The questions must be variety, be definitely inspected from the retinal image, and each question include at least one exam result.
>
> The questions must include a sentence that prompt the answer output in JSON format. Then, the generated answers should be JSON format, for example:
>
> - advanced AMD: 1 if late age-related macular degeneration is present, otherwise 0
>
> - PIG: 1 if abnormal pigmentary is present, otherwise 0
>
> - DRUS: 0 if no drusen or small drusen, 1 if intermediate drusen, 2 if large drusen If the answer includes more than one item, plug a space between
>
> If the answer includes more than one item, plug a space between the two items.
>
> Avoid any associations, interpretations, clinical recommendations and repeating irrelevant information. Here are the patient's basic information and exam results for you to get information about ophthalmology diagnosis.

**Table S4. Example of Generated JSON-form Dialogue.**

"conversations":

[

"from": "human", "value": "<image>Can you tell me if I have intermediate drusen from my exam results? Please answer in JSON format.",

"from": "gpt", "value": "{"DRUS": 1}",

"from": "human", "value": "Do I have advanced age-related macular degeneration based on my retinal image? Please respond in JSON format.",

"from": "gpt", "value": "{"ADVAMD": 0}",

"from": "human", "value": "Is there any abnormal pigmentary in my exam results? Answer in JSON format.",

"from": "gpt", "value": "{"PIG": 0}"

]

**Table S5. Examples of Close-ended Inference Prompt for Evaluation Model Accuracy.**

> You are a highly experienced ophthalmologist specializing in retinal diseases. You will be shown a color fundus photograph of a patient's eye.
>
> Your task is to identify key retinal features and return a structured response. You must only respond in JSON format using the following fields:
>
> - Advanced AMD: 1 if late age-related macular degeneration is present, otherwise 0
>
> - PIG: 1 if pigmentary abnormalities are present, otherwise 0
>
> - DRUS: 0 if no drusen or small drusen, 1 if intermediate drusen, 2 if large drusen
>
> Do not include any explanation, just return the JSON object. Please assess this fundus image and return your findings in the specified JSON format.

**Table S6. Examples of Open-ended Inference Prompt Used in Ophthalmologists Assessment.**

> You are a highly experienced ophthalmologist specializing in retinal diseases.
> You will be shown a color fundus photograph of a patient's eye.
> Your task is to identify key retinal features and return responses.
>
> ADVAMD:
> Are there signs of advanced AMD in this photo?
> Do you think the advanced AMD is present in this image?
> Can you tell me if the advanced age-related macular degeneration is present?
>
> PIG:
> Are there any pigmentary abnormalities in the image?
> Could you tell me if I have any pigmentary abnormalities from the retinal image?
> What are my exam results regarding pigmentary changes?
>
> DRUS:
> What can you tell me about the drusen size and distribution?
> Can you tell me about my drusen size?
> What size are the drusen from my retinal image?
>
> Interpretation:
> Can you be more specific about your findings?
> Can you give me the evidence that support your prediction?
> How do you analyze my retinal image and make this decision, please give me more details?
>
> Additional:
> What else can you tell me about this image?
> Is there anything else I should know?
> Can you detect something else from this image?

**Table S7. Likert Grading Rubric.**

---

**Grading Criteria - Retinal Findings (Core AMD Features)**

**Q1. The model correctly identifies the presence or absence of advanced AMD.**

**Introduction:** The model provides a binary prediction (advanced AMD: Yes or No). Evaluate:

- Is the diagnosis correct based on the fundus image?

- Are hallmark features of advanced AMD (e.g., geographic atrophy, hemorrhage, subretinal fluid, disciform scars) visible to support the diagnosis?

- If asked, does the model give a plausible rationale (even if incomplete or implicit)?

**Scoring Criteria**

• 1: Model prediction is incorrect (e.g., predicts advanced AMD when features are clearly absent, or misses clear signs). No visible features to support claim.

• 2: Prediction may be borderline correct, but lacks clear visible justification (e.g., subtle atrophy or ambiguous signs). Reasoning unclear or absent.

• 3: Prediction is plausible and some image-based evidence is present (e.g., early GA or fibrotic tissue). Minor uncertainty exists.

• 4: Correct diagnosis with supporting visible features. Clinically coherent. Reasoning is plausible if provided.

• 5: Clear and correct prediction with strong visual support (GA or nAMD signs clearly visible). Model would be trustworthy in clinical use.

**Q2. Pigmentary abnormalities are appropriately detected.**

**Introduction:** The model predicts presence/absence of pigmentary abnormalities. Consider:

- Are hypo/hyperpigmented areas present and reasonably interpreted?

- Does the model's output align with what you see?

- If asked, does the model give a plausible explanation (e.g., mentions of mottling, clumping, depigmentation)?

**Scoring Criteria**

• 1: Model missed clear pigmentary changes or falsely claimed them. No visual support. Reasoning (if any) is incorrect.

• 2: Some pigmentary change may be visible, but model prediction is off or ambiguous. Reasoning absent or inaccurate.

• 3: Prediction is plausible with some supporting visual features. Reasoning unclear but not obviously wrong.

• 4: Correctly identifies pigmentary changes with visible support. Reasoning plausible if queried.

• 5: Prediction matches clinical observation; pigmentation abnormalities are clear. Reasoning, if requested, aligns with clinical criteria.

**Q3. Drusen size is accurately categorized.**

**Introduction:** The model outputs a categorical decision: none/small, intermediate, and large. Consider:

- Is the prediction correct based on the image?

- Can you identify visible supporting features for the prediction?

- If you asked the model for an explanation, was the reasoning (if any) correct, incorrect, or unavailable?

**Scoring Criteria**

• 1: Prediction is incorrect, and no clear soft drusen are visible (or falsely claimed). Reasoning (if available) is incorrect or absent.

• 2: Prediction may be correct or borderline, but supporting features are unclear or weak. Reasoning not available or inaccurate.

• 3: Prediction is plausible, and visible evidence (e.g., well-defined, yellowish deposits) is present, even if not explicitly highlighted. Reasoning (if requested) may be absent or partial.

• 4: Correct prediction with supporting image features visible. Reasoning is plausible if provided.

• 5: Correct prediction, supported by clearly visible soft drusen. Reasoning is accurate, or clinician could easily infer the same conclusion independently.

**Q4. Overall Assessment: The model's outputs are useful for clinical decision-making in AMD and general retinal care.**

**Introduction:** Consider:

- Whether the model's diagnostic outputs (e.g., AMD stage, systemic risks) are correct or clinically reasonable.

- Whether it missed any obvious or important findings visible in the image.

- Whether it hallucinated findings (i.e., reported conditions do not present in the image).

**Scoring Criteria**

• 1: Outputs are mostly incorrect, miss key findings, or include hallucinated diagnoses. Clinical utility is low or potentially misleading.

• 2: Some outputs are reasonable, but the model either misses important features or makes unsupported claims. Use would require significant clinician correction.

• 3: Outputs are partly useful — major findings generally present, some minor omissions or questionable statements. Would use with caution.

• 4: Most outputs are correct and relevant. Only minor omissions or borderline findings; enhances efficiency with limited oversight.

• 5: Diagnoses are accurate, important findings are captured, and no hallucinations observed. Highly useful in supporting clinical decisions.